\documentclass{article}
\usepackage[numbers, compress]{natbib}

\usepackage[final]{neurips_2023_gaied}
\usepackage{graphicx}
\usepackage{amsmath}
\usepackage{multirow}
\usepackage{amssymb}
\usepackage{caption}
\usepackage{longtable}
\usepackage{pifont}

\usepackage[utf8]{inputenc} 
\usepackage[T1]{fontenc}    
\usepackage{hyperref}       
\usepackage{url}            
\usepackage{booktabs}       
\usepackage{amsfonts}       
\usepackage{nicefrac}       
\usepackage{microtype}      
\usepackage{xcolor}         

\title{Automated Distractor and Feedback Generation for \\Math Multiple-choice Questions via \\In-context Learning}

\author{%
  Hunter McNichols$^*$, Wanyong Feng$^*$, Jaewook Lee$^*$, Alexander Scarlatos \\
  University of Massachusetts Amherst \\
  \texttt{\{wmcnichols,wanyongfeng,jaewooklee,ajscarlatos\}@umass.edu} \\
  \AND
  Digory Smith, Simon Woodhead \\
  Eedi \\
  \And
  Andrew Lan \\
  University of Massachusetts Amherst\\
}

\begin{document}

\maketitle
\def\thefootnote{*}\footnotetext{These authors contributed equally to this work}

\begin{abstract}
Multiple-choice questions (MCQs) are ubiquitous in almost all levels of education since they are easy to administer, grade, and are a reliable form of assessment. An important aspect of MCQs is the distractors, i.e., incorrect options that are designed to target specific misconceptions or insufficient knowledge among students. To date, the task of crafting high-quality distractors has largely remained a labor-intensive process for teachers and learning content designers, which has limited scalability. In this work, we explore the task of automated distractor and corresponding feedback message generation in math MCQs using large language models. We establish a formulation of these two tasks and propose a simple, in-context learning-based solution. 
%
%
Moreover, we propose generative AI-based metrics for evaluating the quality of the feedback messages. 
%
%
We conduct extensive experiments on these tasks using a real-world MCQ dataset. 
Our findings suggest that there is a lot of room for improvement in automated distractor and feedback generation; based on these findings, we outline several directions for future work. 
\end{abstract}

\section{Introduction}
\begin{figure*}[t]
\centering
\includegraphics[width=1\linewidth]{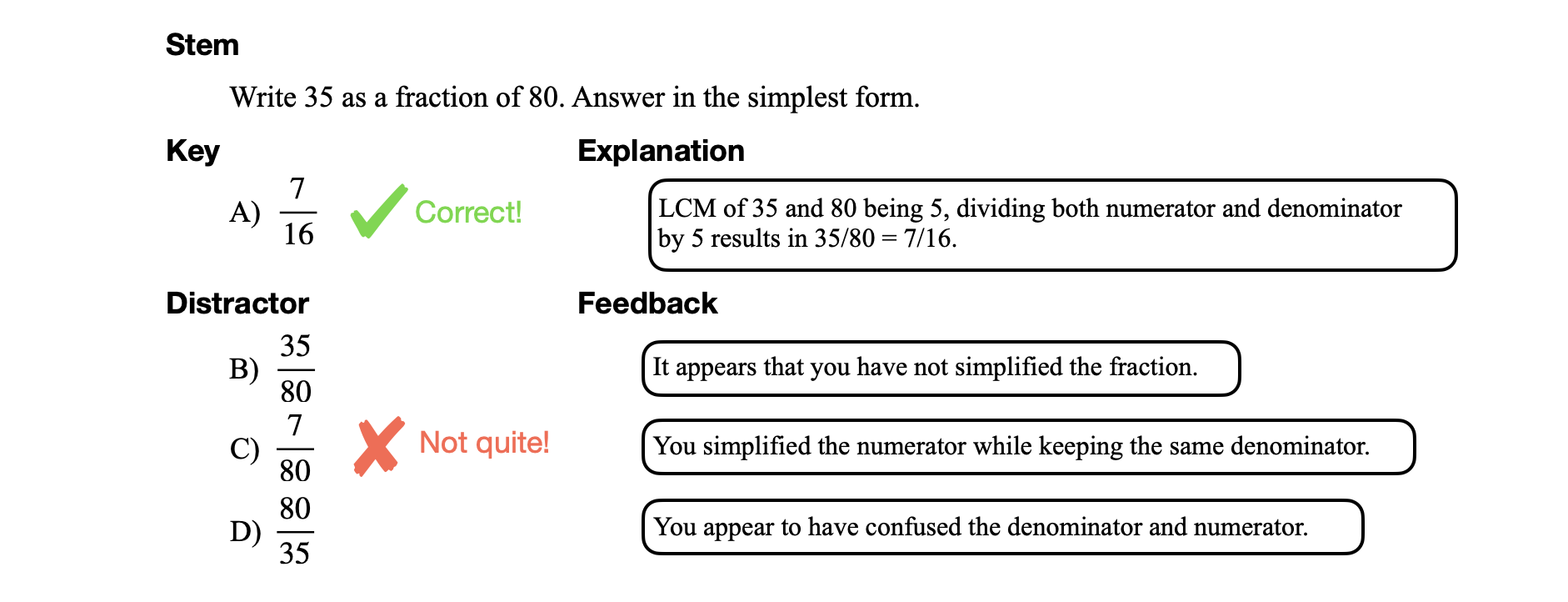}
\caption{Different parts of math MCQs illustrated with an example.}
\label{fig:terminology}
\end{figure*}

Multiple-choice questions (MCQs) are widely used to evaluate students knowledge since they enable quick and accurate administration and grading. They are reliable since they are designed to measure specific learning objectives consistently~\cite{Nitko:96, Airasian:01, Kubiszyn:16}. MCQs are constructed in a specific format; see Figure~\ref{fig:terminology} for an example. The \textit{stem} refers to the statement on the problem setup and context, followed by a question that needs to be answered. Among the options, the correct one can be referred to as the \textit{key}, while incorrect ones can be referred to as \textit{distractors}. As the name implies, distractors in multiple-choice questions are typically formulated to align with common mistakes made by students. These distractors are chosen because students either i) lack the necessary comprehension of the concepts or skills tested in the question to accurately identify the key as the correct answer, or ii) hold misconceptions that result in selecting a specific distractor as their response.

While MCQs offer many advantages for evaluating students' knowledge, manually crafting high-quality MCQs is a demanding and labor-intensive process~\cite{kelly2013traditional}. High-quality distractors should be plausible enough to mislead students but not so evidently incorrect as to be easily discernible. In addition, each distractor should include a feedback message to aide the student in identifying their error or misconception and lead them to the correct answer. Therefore, crafting ideal distractors and feedback is a challenging task which requires a deep understanding of possible student misconceptions.

Prior work on automatic distractor generation primarily focused on language learning and reading comprehension tasks, where distractors are used to assess students' comprehension of a given text or article. Early works used a ranking method based on semantic similarity and word collocation information or a pre-defined ontology to produce distractors~\cite{susanti2018automatic, stasaski2017multiple}. More recent works used encoder-decoder models with attention mechanisms for distractor generation, resulting in longer and higher-quality distractors~\cite {qiu2020automatic, shuai2023qdg, xie2021diverse, gao2019generating}. Additionally, several recent works used pre-trained LLMs such as BERT and T5 for distractor generation in the context of Swedish reading and Cloze test~\cite{kalpakchi2021bert, chiang2022cdgp, rodriguez2022end}. However, to the best of our knowledge, there is limited work on distractor generation for math MCQs. Existing works either used constraint logic programming or sent predefined errors to a math problem solver to generate distractors~\cite{tomas2013automatic, dave2021math}.
Prior work on feedback generation primarily focuses on language learning \cite{speech, foreign-language} and programming \cite{feedback-gen-programming-intro, feedback-gen-programming-intro-performance, feedback-gen-programming-context, feedback-gen-programming-grading, feedback-gen-programming-context-novice, jia2022insta} settings. These works typically leverage template-based rules or, occasionally, fine-tuned language models. For math, it is possible to use teachersourcing \cite{patikorn2020effectiveness} or cognitive models \cite{Koedinger:06, liu2016cogn, orourke2019algebra, ritter2007cogntutor} to anticipate common student errors and deploy pre-written feedback messages associated with those errors \cite{lan-grading-and-feedback, zhang2021math}, usually in intelligent tutoring systems \cite{feedback-effect-math, feedback-effect-its}. However, to the best of our knowledge, there is no prior work on automatically generating math feedback using LLMs or generating math feedback for MCQs. 

\subsection{Contributions}
In this work, we investigate the tasks of automatically generating distractors and corresponding feedback messages for MCQs in the domain of math problems, in particular with LLMs. The first task is to generate plausible and challenging distractors for math MCQs automatically, given the question stem and (optionally) the key and its corresponding explanation. The second task is to generate informative and instructive feedback messages for each distractor automatically, given the question stem and the specific distractor. We believe that these two tasks can potentially be crucial to scaling up the design and support of MCQs in assessment and learning scenarios. We study a simple approach for the distractor and feedback generation tasks using few-shot, in-context learning~\cite{brown2020fewshot} where the in-context examples are chosen as other MCQs with the highest semantic similarity, an approach that can serve as a baseline for future work.
%
%
To evaluate the effectiveness of feedback, we explore using two \textit{reference-free} metrics that leverage generative LLMs to evaluate the helpfulness and informativeness of the feedback. We discuss the potential benefits and limitations of these metrics compared to standard text similarity-based metrics in detail. 
 We conduct extensive experiments on a real-world dataset of math MCQs with student response statistics, analyze our results both quantitatively and qualitatively, and suggest directions for future work. 

\begin{figure*}[t]
\centering
\includegraphics[width=.9\linewidth]{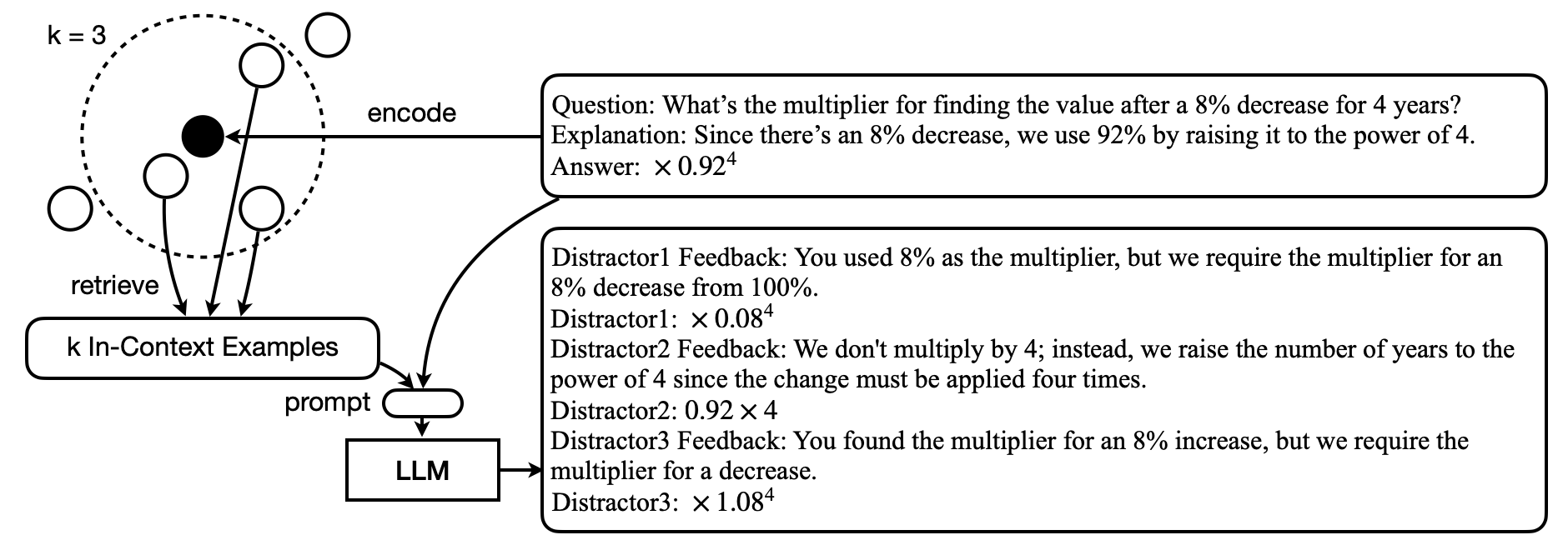}
\vspace{-.2cm}
\caption{Overview of distractor generation with a math MCQ on ``compound percentage decrease''.}
\label{fig:knn_example}
\end{figure*}

\section{Proposed Tasks}
\label{sec:proposed_task}
In this section, we first define relevant mathematical notation in MCQs and then formally define the automated distractor and feedback generation tasks. We define an MCQ $Q$ as a set of textual components defined below. In this paper, we do not consider questions that contain diagrams or images; extending our work to multi-modal question content is left for future work.
\begin{align*}
    Q = \{s, k, e_k, D, F\}.
\end{align*}
Here, $Q$ contains a stem $s$, a key $k$, and an (optional) explanation of the key $e_k$. All of these components are sequences of words and math symbols (e.g., $s=\{w_1, \dots, w_L\}$ where $L$ is the length of the sequence $s$). In addition, $Q$ contains a set of distractors $D$, each of which has an (optional) corresponding feedback message $f_i$ which is shown to a student upon selecting $d_i$. The set of feedback messages is denoted as $F$. The distractors and feedback messages are likewise sequences of words and math symbols. See Figure~\ref{fig:terminology} for an example MCQ.

\subsection{Distractor Generation}
We formulate the task of distractor generation as learning a function $g^\text{dis}$ that outputs a set of distractors $\hat{D}$ for an MCQ given the question stem and (optionally) the key and its explanation, i.e.,
\begin{align*}
g^\text{dis}(s, k, e_k) \rightarrow \hat{D}
\end{align*} 
Our goal is to generate distractor options that students with specific misconceptions or insufficient knowledge on skills required for the question will select. This way, the MCQ can better distinguish between students that master all of the skills required for the question and those that do not. See Section~\ref{sec:eval_metrics} for details on our evaluation metrics for generated distractors.

\subsection{Feedback Generation}

We formulate the task of automated feedback generation as learning a function $g^\text{fb}$ that outputs a textual feedback message given the question stem, a distractor, and (optionally) the key $k$ and its explanation $e_k$. Note that in our setting, we generate a single feedback message that corresponds to each distractor, i.e.,
\begin{align*}
g^\text{fb}(s, d_i, k, e_k) \rightarrow f_i.
\end{align*}
Our goal is to generate feedback messages for students who select a distractor to enhance their understanding of why their answer is incorrect and guiding them towards correcting their error.

\section{Methodology}
In this section, we detail our preliminary approach towards solving the tasks formulated in Section~\ref{sec:proposed_task}. We then define the metrics we use to evaluate our approaches to each task.

\subsection{In-Context Learning}
Our approach to solving both distractor and feedback generation tasks is to use a LLM to estimate the functions $g_\phi^\text{fb}$ and $g_\phi^\text{dis}$, where $\phi$ are LLM parameters.  We utilize in-context learning with similar MCQs chosen by the k-nearest neighbor (\textbf{kNN}) algorithm as few-shot examples for LLM input, since kNN has been shown to be an effective example selection heuristic for LLM prompting ~\cite{incontext2021gpt3}. Specifically, given a training dataset of MCQs, we first partition the dataset into two subsets: an example candidate set and a test set. To generate distractors for each target question in the test set, we employ a kNN approach to select a predetermined number of example questions from the candidate set. These selected examples are subsequently integrated as in-context samples in our prompts for the LLM. Figure~\ref{fig:knn_example} provides a visual representation of this process.
To determine similarity, we calculate the \textit{cosine similarity} between vectorized textual encodings of the examples and the target MCQ. We obtain these encodings using a pre-trained SBERT encoder \texttt{mpnet}~\cite{reimers2019sentencebert}, to which we provide a concatenated string of all MCQ components as input. See Table~\ref{tab:prompt_format} for details on our prompt format.

\subsection{Evaluation Metrics}
\label{sec:eval_metrics}

\subsubsection{Distractor Generation}



We utilize alignment-based metrics to evaluate the extent to which the generated distractors align with the ground-truth, human-authored distractors. We denote the generated distractors as $\hat{D}$ where $|\hat{D}|=N$, the number of distractors in the MCQ. Prior to computing these metrics, the predicted and ground truth sets must be index-aligned. We introduce two binary metrics \textbf{Partial} match $h_p$ (whether at least one generated distractor matches a ground-truth one) and \textbf{Exact} match $h_e$ (whether all generated distractors match ground-truth ones), defined formally as:

\begin{minipage}{.5\linewidth}
   \begin{align*}
    h_p(D,\hat{D}) = \begin{cases} 
      1 & \exists{\hat{d_i}} \in{\hat{D}} : \hat{d_i} = d_i \\
      0 & \text{otherwise}
    \end{cases}
\end{align*} 
\end{minipage}%
and
\begin{minipage}{.5\linewidth}
    \begin{align*}
    h_e(D,\hat{D}) = \begin{cases} 
        1 & \forall{\hat{d_i}} \in{\hat{D}} : \hat{d_i} = d_i \\
        0 & \text{otherwise}.
    \end{cases}
\end{align*}
\end{minipage}
We also utilize a continuous measure in the range $[0,1]$ that we call \textbf{Proportional} match $h_n$ (the portion of generated distractors that match ground-truth ones), i.e., 
\begin{align*}
    h_n(D, \hat{D}) = \frac{\sum_{i=1} \mathbf{1}_{i: \hat{d}_i = d_i}}{N}
\end{align*}
where $\mathbf{1}$ denotes an indicator function. We report metrics in Table~\ref{tab:methodology_comparison} by averaging across all target MCQs in the test set and scale the values by a factor of 100 into percentages.

\par
\subsubsection{Feedback Generation} 
We utilize both existing reference-based and novel reference-free metrics to evaluate generated feedback messages. For reference-based metrics, we compare the actual, ground-truth human-written feedback for a distractor with the generated one using \textbf{BLEU} \cite{bleu}, a precision-based metric, \textbf{ROUGE-L} \cite{rouge}, a recall-based metric, and \textbf{METEOR} \cite{meteor}, a metric that leverages precision, recall, and word order. However, these metrics are imperfect since they do not target the underlying goal of feedback generation: to be informative to a student and help them correct their errors. Therefore, these metrics are susceptible to false positives (when the generated feedback matches well with the reference structurally but has key factual/logic errors) and false negatives (when the generated feedback does not match the reference structurally but is still valid and helpful). While it is common to use human evaluation \cite{jia2022insta} as a high-quality alternative to these metrics, it can be expensive and time-consuming. As an alternative, we propose two automated reference-free metrics, where we employ a generative LLM to act as a surrogate for human evaluators, along the lines of \cite{feedback-eval-p2p,feedback-eval-zero}. We design these metrics based on the observation that feedback messages tend to play two primary roles: to be \textit{helpful} for students to reach a correct answer, and to be an \textit{explanation} for why an answer is incorrect.

\paragraph{Answer Adjustment} This metric measures if a feedback message is \textit{helpful} for correcting the corresponding error in a distractor. Specifically, we prompt a generative LLM to take steps indicated by the feedback to make an adjustment to the distractor, or formally, $g^\text{adj}(s, d_i, f_i) \rightarrow k$. We then measure if the adjusted answer is correct, indicating that the feedback is helpful, or incorrect, indicating that the feedback is misleading or not helpful. When reporting results, we denote the percentage of samples where the LLM adjusts the distractor to the correct answer as \textbf{Adj.}

\paragraph{Distractor Prediction}
This metric measures if a feedback message serves as an \textit{explanation} for why a distractor is incorrect. Specifically, we prompt a generative LLM to use the feedback in order to predict what the corresponding distractor is, or formally, $g^\text{dp}(s, k, f_i) \rightarrow d_i$. We then measure if the predicted answer is equal to the distractor, indicating that the feedback is informative, or not equal to the distractor, indicating that the feedback is misleading or too generic. When reporting results, we denote the percentage of samples where the LLM correctly predicts the distractor as \textbf{Dist. Pred.}

\section{Experiments}

\subsection{Dataset, Methods, Experimental Setup}

Our dataset consists of 1.4K MCQs from Eedi's \footnote{\url{https://eedi.com/home}} content repository, after filtering out questions with images/diagrams. Each question has four answer options, of which only one is correct, while the remaining options serve as distractors intended to reveal student misconceptions. See Figure~\ref{fig:terminology} for an example of the different components in a MCQ, including explanation for the key and feedback for distractors. The questions are sourced from the broad mathematical topic titled ``Number'' with subtopics including ``Basic Arithmetic'', ``Fractions'', and ``Rounding and Estimating''. The questions are primarily targeted towards students aged 10-13. 
%
%
We divided the dataset into two subsets, namely a candidate example set and a test set, using an 80:20 ratio. The candidate example set is utilized to select MCQs as in-context examples and the test set contains all the target questions for generation and analysis. Across our experiments, we use a variety of prompt formats and corresponding encoding strategies; these formats are summarized in Table~\ref{tab:prompt_format} in the Supplementary Material. 
For example, $\text{kNN}^{\text{all}}$ means that i) we use both the key and explanation in addition to the question stem to calculate similarities and ii) we use them together with all distractors and feedback messages of each example MCQ as input to the LLM. 

For baselines, we include a \textbf{Random} heuristic that randomly selects few-shot examples for in-context learning. The prompt format for Random follows that of $\text{kNN}^{\text{all}}$, our best performing prompt format. We use three in-context examples for all few-shot methods. We also include \textbf{Zero-shot} as an additional baseline, where a generative LLM is solely provided with detailed instructions and information on the distractor and feedback generation tasks, without any in-context examples. We detail our prompts in the Supplementary Material, Section~\ref{sec:prompts}. 
We emphasize that there is no rule-based baeline that we can compare against since the MCQs in our real-world dataset are created by math teachers, not from a set of rigid rules. 
We use the OpenAI models \texttt{Codex} (code-davinci-002) \cite{codex} and \texttt{ChatGPT} (gpt-3.5-turbo) \cite{chatgpt} as our generative LLMs, with greedy decoding and a maximum output length of 256 tokens. We employ 
\texttt{Codex} for both kNN and Random due to its proficient comprehension of mathematical content and ability in few-shot learning. We use the larger \texttt{ChatGPT} and \texttt{GPT-4} models for Zero-shot. 
We use \texttt{ChatGPT} for the reference-free feedback evaluation metrics; please see the Supplementary Material for all prompts we used in our experiments. 


\subsection{Results and Discussion}

\begin{table}
    \begin{minipage}{.4\linewidth}
        \scalebox{0.75}{
            \begin{tabular}{lcccc}
            \hline
            \textbf{Method}  & Exact   & Partial   & Proportional     \\\hline
            $\text{kNN}^{\text{all}}$  & \textbf{10.06}   & \textbf{71.02}     & \textbf{38.16} \\
            $\text{kNN}^{\text{none}}$ & 6.01    & 54.52     & 27.20    \\
            $\text{kNN}^{\text{key}}$  & 8.13    & 61.48     & 32.39  \\\hline
            Random     & 1.77    & 52.30     & 22.85      \\
            $\text{Zero-shot}^{\text{ChatGPT}}$     & 1.77    & 50.09     & 21.79          \\
            $\text{Zero-shot}^{\text{GPT-4}}$     & 3.18    & 44.52    & 21.67      \\\hline
            $\text{kNN}^{\text{all}}_{\neg T}$    & 3.89    & 55.83     & 25.91   \\
            \end{tabular}    
            }
            \vspace{0.05\linewidth}
        \caption{Results of distractor generation where kNN-based methods often significantly outperform baselines.}
        \label{tab:methodology_comparison}
    \end{minipage}
    \hspace{0.1cm}
        \begin{minipage}{.54\linewidth}
    \scalebox{0.7}{
    \begin{tabular}{lccccc}
    \hline
    {\bfseries Method} &  {BLEU} & {ROUGE-L} & {METEOR} & {Adj.} & {Dist. Pred.} \\
    \hline
    Ground-truth & -- & -- & -- & \textbf{49.00} & \textbf{24.73} \\
    $\text{kNN}^{\text{one}}$ & \textbf{33.70} & \textbf{42.28} & \textbf{43.64} & 46.64 & 18.26 \\
    $\text{kNN}^{\text{one}}_{\neg T}$ & 13.04 & 25.65 & 26.83 & 42.05 & 15.55 \\
    Random & 4.21 & 20.08 & 18.63 & 42.17 & 13.19 \\
    Zero-shot & 3.12 & 17.62 & 18.05 & \underline{47.70} & \underline{20.49} \\
    \end{tabular}
    }
    \vspace{0.2cm}
    \caption{Evaluation of generated feedback messages on reference-based and reference-free metrics.}
    \label{table:feedback-gen}
    \end{minipage}
\end{table}

\paragraph{Distractor Generation}
Table~\ref{tab:methodology_comparison} shows results on distractor generation. We experiment with kNN across three prompt formats together with Random and Zero-shot prompting. To test generalizability, we also experiment with restricting the candidate example set by excluding questions from the same sub-topic. 
%
%
We observe that kNN-based methods have much higher performance on the alignment-based metrics compared to the Random and Zero-shot baselines.
%
%
This result is not surprising since in some cases the target and example questions are highly similar and only differ in numerical values. Therefore, the LLM can generate distractors that match the ground-truth ones by simply replicating the style of the in-context samples and substituting the correct numbers. The baselines score significantly lower on these metrics, as the methods do not find or have informative in-context samples. Moreover, we observe that $\text{kNN}^{\text{all}}$ surpasses other kNN-based methods, as it uses the most information to identify in-context samples that are similar to the target question. 


Next, we detail the impact of removing the same sub-topic from the candidate example set with $\text{kNN}^{\text{all}}_{\neg T}$. As shown in Table~\ref{tab:methodology_comparison}, doing so results in a huge performance drop-off from $\text{kNN}^{\text{all}}$. To further illustrate this drop-off, we perform a case study of the question described in Figure~\ref{fig:knn_example}. To answer the MCQ in the Figure correctly, a student should understand and apply two math concepts: percentage decrease and compound interest. Ideally, good distractors should be designed to embed anticipated errors in understanding for both concepts needed to answer the question. 
For this question, $\text{kNN}^{\text{all}}$ generates all ground-truth distractors correctly, with the exact same anticipated errors as the ground-truth.  Two distractors anticipate the error of misinterpreting ``decrease by 8\%'' as either an increase by 8\% or simply focusing on the 8\% number itself. The other one anticipates an error of confusing compound interest with simple interest. Examining the in-context examples, we found that one retrieved example has an identical format to the question (differing only in numerical values), which helps the LLM generate the ground-truth distractors. The exact in-context examples retrieved are shown in Table~\ref{table:distractor-kNN-SEKFD}.
However, $\text{kNN}^{\text{all}}_{\neg T}$ only generates one of the correct distractors, $\times1.08^4$. The other two non-matching distractors $\times1.08$ and $\times1.08^2$ are not related to misconceptions in compound interest. This result can be explained by examining the in-context examples retrieved by $\text{kNN}^{\text{all}}_{\neg T}$. The exact in-context examples retrieved are shown in Table~\ref{table:distractor-kNN-SEKFD*}. Since questions with the same sub-topic are removed from the candidate example set, $\text{kNN}^{\text{all}}_{\neg T}$ chooses questions that ask the following: i) multiplier for decreasing by a certain amount and ii) years to reach a certain value with depreciation. Since kNN relies on semantic similarity, it focuses only on the concept of percentage decrease, failing to find in-context examples of compound interest, which results in low-quality distractors. This case study highlights a major limitation in using semantic similarity as a proximity measure for kNN-based methods, which we leave for future work. 

\paragraph{Qualitative Analysis for Distractor Generation}
\label{sec:qual}

We now qualitatively investigate the distractors generated by the best performing setting, $\text{kNN}^{\text{all}}$, to extract some insights on the distractor generation task and how to improve performance. We group questions into 4 categories according the number of generated distractors that match the human-authored ones, from 0 to 3.

For the group where all generated distractors match the ground truth (3 out of 3), we find that, in all but 2 of the 30 cases, there was an in-context example that is very similar to the target question, with the only difference being different numerical values or names. However, this situation often appears in other categories as well, which is perhaps surprising since it implies that the presence of a near-identical in-context example alone is not sufficient for an LLM to generate quality distractors. Another possible explanation is that for two near-identical questions, their sets of distractors may differ, which suggests that the number of appropriate distractors for each question differs and there is a lot of variance in which set of 3 is actually used. 

For the group where none of the generated distractors match the ground truth, we examine 22 of the 82 cases. We find that in 17 of the 22 cases (77\%), the generated distractors are reasonable and the ground truth distractors are not clearly superior to the generated ones; see Table~\ref{table:distractor-example-mulitples} for an example. While this observation is entirely subjective, it highlights that alignment-based metrics may not be an appropriate metric to measure the quality of distractor generation since human-authored distractors may not be optimal. 
%
%
Moreover, since many generated distractors are actually reasonable even if they are not among the human-authored ones, there is promise in using automated distractor generation for teacher support during the creation of MCQs. 

Finally, for the groups where 1 or 2 generated distractors match the ground truth, we examine which human-authored distractor(s) were generated and which were not. We find that in many cases, one of the generated distractors that matches the ground truth seems to come from a typical misconception relevant to the question, while other human-authored distractors do not seem to correspond to any typical misconception; see Table~\ref{table:distractor-example-fraction-simplify} for an example.


%

\paragraph{Feedback Generation}


Table~\ref{table:feedback-gen} shows results on feedback generation. We experiment with $\text{kNN}^{\text{one}}$, Random, and Zero-shot prompting, as well as kNN prompting with questions in the same topic excluded from the pool ($\text{kNN}^{\text{one}}_{\neg T}$). As a reference, we also show results of the ground-truth feedback messages on the reference-free metrics. We observe that $\text{kNN}^{\text{one}}$ has the best performance on the reference-based metrics by far compared to other methods. This result is likely due to many feedback messages being almost identical for similar distractors across questions in the same topic; as a result, the LLM can generate feedback messages that are highly similar to the ground message by simply copying the style of the in-context examples. The remaining methods score much lower on these metrics, mainly because they are not able to copy in the same way. However, the evaluation is much closer across methods on the reference-free metrics, which indicates that the reference-based metrics are overly biased with respect to the style of the ground-truth feedback.
The ground-truth feedback messages outperform all prompting methods on the Adj. and Dist. Pred. metrics, as expected. Among the prompting methods, Zero-shot performs the best, although we observe that the messages generated with this method are more likely to include the correct answer or distractor in the feedback itself, making it easier to perform well on the metrics without needing high quality feedback. We also observe that $\text{kNN}^{\text{one}}$ outperforms the remaining methods on these metrics and that it is less likely than Zero-shot to copy the answer or distractor into the feedback. We provide an example of feedback generated by each method on a single question/distractor in Table~\ref{table:feedback-gen-across-methods}.

There is a relatively small difference between the generation methods and the ground-truth feedback on the reference-free metrics. This observation is surprising since we expect the ground-truth feedback to be significantly better than most generated feedback messages. It is also surprising that the LLM fails to correct the answer even using the ground-truth feedback since recent models like \texttt{ChatGPT} are much more capable in math reasoning compared to earlier models. We study several common cases for \textit{false negatives} and \textit{false positives} to explain this result. 

For false negatives, many questions are framed in the context of all options, such as asking the student to select the best estimate among options or select the option that satisfies a particular constraint. This setting makes the evaluation tasks difficult since each feedback message is evaluated in the context of a single distractor. Moreover, we observe that the LLM often ignores the instructions on predicting a distractor and simply outputs the correct answer. Furthermore, we observe that the LLM often misinterprets the meaning of the feedback, even if it is indeed high-quality, leading to incorrect predictions. 

For false positives, the LLM often correctly solves the problem on its own, even if the given feedback is unhelpful or inaccurate. Moreover, the LLM often makes a lucky guess of the distractor even if the feedback is not specific enough. Overall, we believe that most of these issues are due to limitations of the LLM and could likely be relieved with further prompt engineering or by using more capable models such as \texttt{GPT-4}, which we leave for future work. 

Despite these limitations, we argue that these metrics provide meaningful results on the \textit{relative} performance of methods compared to each other and are significantly less biased than the reference-based metrics. For more details, we provide instances where the metrics work well in Tables \ref{table:feedback-eval-success-pass}, \ref{table:feedback-eval-success-adj-fail}, and \ref{table:feedback-eval-success-dp-fail}, as well as an instance where the metrics fail in Table \ref{table:feedback-eval-fail}, in the supplementary material.

\section{Conclusions, Future Work, and Acknowledgement}

In this paper, we have formulated two tasks for multiple-choice questions: automated distractor and feedback generation. We experimentally validated large language model-based approaches for these tasks on a real-world dataset. We reported our results using a series of metrics, including standard ones and two novel ones we proposed. The results have shown that these tasks are still challenging despite rapid recent advancement in large language models. 
Our initial exploration opens up many avenues for future work. 
%
%
First, we need to explore approaches other than large language model prompting such as fine-tuning a smaller, task-specific model (although we did not find fine-tuning to be effective in some preliminary experiments). Second, we need to develop modified text encoding methods that are more closely aligned with student errors, which will enable us to move away from using surface semantic similarity in question statement to select in-context examples. Finally, we need to conduct a human evaluation on the generated distractors and feedback messages. 
The authors thank the NSF (under grants 2118706 \& 2237676) and Schmidt Futures for supporting this work.

\newpage
\bibliographystyle{unsrt}
\bibliography{custom}
\newpage

\section*{Supplementary Material}

\begin{table*}[tp]
\centering
        \scalebox{0.9}{
            \begin{tabular}{lccc|cccc}
            \hline
            \multirow{2}{*}{\bfseries{Method}} & \multicolumn{3}{c|}{Encode} & \multicolumn{4}{c}{Prompt} \\\
                                    & $k$    & $e_k$    & $d$    & $k$  & $e_k$ & $F(f)$ & $D(d)$ \\\hline
            $\text{kNN}^{\text{none}}$                   & no      & no     & no      & no    & no  & none  & all  \\
            $\text{kNN}^{\text{key}}$                   & yes      & no    & no      & yes    & no & none  & all  \\
            $\text{kNN}^{\text{all}}$                  & yes     & yes    & no      & yes   & yes & all   & all  \\
            $\text{kNN}^{\text{one}}$                & no      & no     & yes     & no    & no  & one   & one 
            \end{tabular}        
        }
        \vspace{0.1cm}
        \caption{Encoding strategies for retrieval and prompt formats used in kNN-based methods.}
        \label{tab:prompt_format}
\end{table*}

\subsection{Prompts}
\label{sec:prompts}

We provide the prompts we use for several zero-shot tasks in the work below. We use $<>$ to indicate that a variable is filled in dynamically.

\begin{minipage}{0.9\textwidth}
    \strut\newline
    \centering
    \includegraphics[width=.9\linewidth]{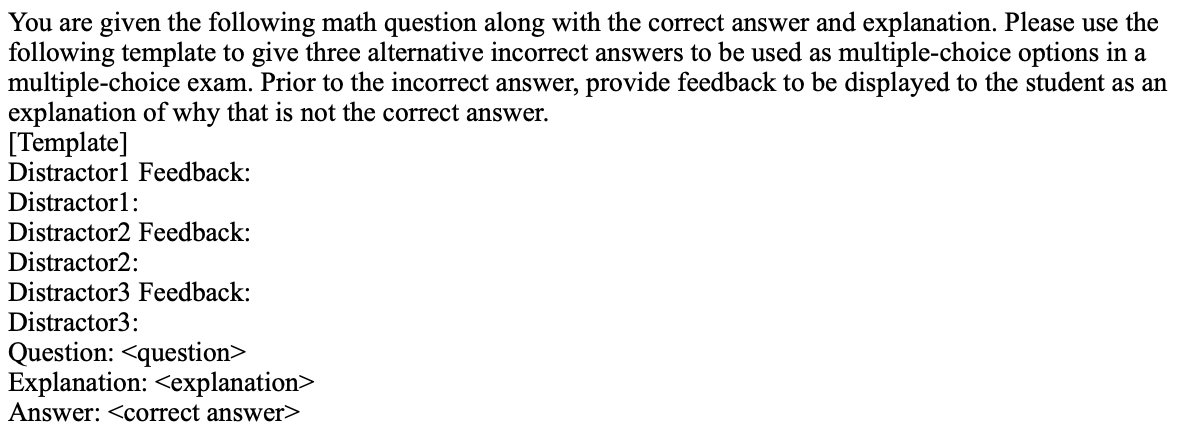}
    \captionof{figure}{Distractor generation zero-shot prompt.}
    \label{fig:prompt_distractors_zero_shot}
\end{minipage}

\begin{minipage}{0.9\textwidth}
    \strut\newline
    \centering
    \includegraphics[width=.9\linewidth]{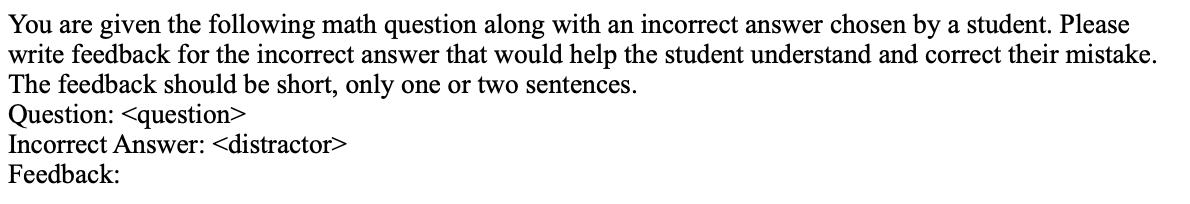}
    \captionof{figure}{Feedback generation zero-shot prompt.}
    \label{fig:prompt_feedback_zero_shot}
\end{minipage}

\begin{minipage}{0.9\textwidth}
    \strut\newline
    \centering
    \includegraphics[width=.9\linewidth]{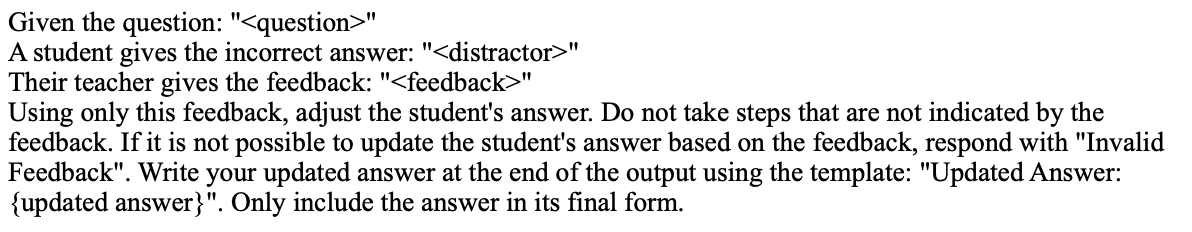}
    \captionof{figure}{Answer adjustment evaluation prompt.}
    \label{fig:prompt_adjust_eval}
\end{minipage}

\begin{minipage}{0.9\textwidth}
    \strut\newline
    \centering
    \includegraphics[width=.9\linewidth]{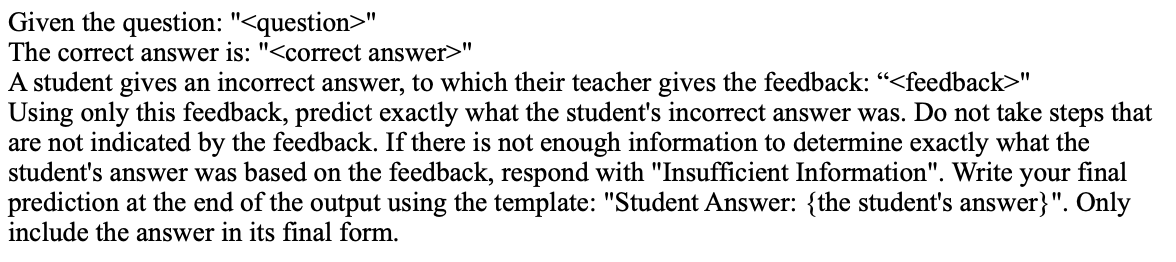}
    \captionof{figure}{Distractor prediction evaluation prompt.}
    \label{fig:prompt_dist_pred_eval}
\end{minipage}

\clearpage

\subsection{Examples of Generated Feedback}
\label{sec:feedback-examples}

\begin{minipage}{\linewidth}
    \strut\newline
    \begin{tabular}{p{5.5cm}p{8cm}}
        \toprule
        \textbf{Question} & which multiplier can be used to find the value after an amount has decreased in value by $8 \%$ for 4 years? \\
        \midrule
        \textbf{Correct Answer} & $\times 0.92^{4}$\\
        \midrule
        \textbf{Distractor} & $\times 1.08^{4}$\\
        \midrule
        \textbf{Ground-truth Feedback} & It looks like you have found the multiplier for an increase of 8\% but we need to find the decrease.\\
        \midrule
        \textbf{Generated Feedback ($\text{kNN}^{\text{one}}$)} & It looks like you have found the multiplier for an increase of 8\% but we need to find the decrease. \\
        \midrule
        \textbf{Generated Feedback ($\text{kNN}^{\text{one}}_{\neg T}$)} & This is the multiplier for increasing by 8\% for 4 years. We are decreasing by 8\% for 4 years. \\
        \midrule
        \textbf{Generated (Random)} & It looks like you have found the multiplier for a decrease of 8\% for 1 year. \\
        \midrule
        \textbf{Generated Feedback (Zero-shot)} & The incorrect answer uses the wrong multiplier. To find the value after a decrease, you need to use a multiplier less than 1. \\
        \bottomrule
    \end{tabular}
    \captionof{table}{Examples of feedback generation across prompting methods. $\text{kNN}^{\text{one}}$ generates the exact same message as the ground-truth feedback, $\text{kNN}^{\text{one}}_{\neg T}$ generates a similar message that is still valid, Random generates a similar message that is invalid, and Zero-shot generates a dissimilar message that is still valid though less informative.}
    \label{table:feedback-gen-across-methods}
\end{minipage}

\begin{table*}[hb]
    \begin{tabular}{p{4.5cm}p{9cm}}
        \toprule
        \textbf{Question} & sheila is listing the multiples of 7 in ascending order. she start with 7. what number should she write next? \\
        \midrule
        \textbf{Correct Answer} & 14\\
        \midrule
        \textbf{Distractor} & 1\\
        \midrule
        \textbf{Ground-truth Feedback} & Not quite, I think you divided by 7. We need to find the multiples of 7, the numbers in the 7 times table\\
        \textbf{Adjustment Prediction} & 14 \color{green}\ding{51}\\
        \textbf{Distractor Prediction} & 1 \color{green}\ding{51}\\
        \midrule
        \textbf{Generated Feedback ($\text{kNN}^{\text{one}}$)} & I think you're confusing factors and multiples; 1 is the highest common factor of 7 and 7. The word 'multiple', has more of a clue to it ... think 'multiply', so the times table for that number.\\
        \textbf{Adjustment Prediction} & 14 \color{green}\ding{51}\\
        \textbf{Distractor Prediction} & 1 \color{green}\ding{51}\\
        \midrule
        \textbf{BLEU} & 0.0 \\
        \textbf{ROUGE-L} & 33.33 \\
        \textbf{METEOR} & 35.53 \\
        \bottomrule
    \end{tabular}
    \caption{Example of feedback evaluation, for both ground-truth feedback and feedback generated by $\text{kNN}^{\text{one}}$ prompting. Generated predictions for both reference-free metrics are shown (with generated reasoning steps removed for brevity). Both feedback messages are high quality and pass the reference-free metrics.}
    \label{table:feedback-eval-success-pass}
\end{table*}

\begin{table*}[hb]
    \begin{tabular}{p{4.5cm}p{9cm}}
        \toprule
        \textbf{Question} & a 3 digit number belongs in this box to make the calculation true \[ 467+\square=1253 \] what is the 3rd digit of the number that belongs in the box? \\
        \midrule
        \textbf{Correct Answer} & 6\\
        \midrule
        \textbf{Distractor} & 4\\
        \midrule
        \textbf{Ground-truth Feedback} & I think you've found what would be added to 3 to get 7. Adding 4 in the units would give an answer with units of 1. Here we need a number to add to 7 to get to 3 (or 13)\\
        \textbf{Adjustment Prediction} & 6 \color{green}\ding{51}\\
        \textbf{Distractor Prediction} & 4 \color{green}\ding{51}\\
        \midrule
        \textbf{Generated Feedback ($\text{kNN}^{\text{one}}$)} & 4 is the units in the final value, however it is not what we have to add to 467. Adding 4 in the units column would give us 7 units in the answer. What can we add to 7 to get 4 (or 14)?\\
        \textbf{Adjustment Prediction} & 7 \color{red}\ding{55}\\
        \textbf{Distractor Prediction} & 4 \color{green}\ding{51}\\
        \midrule
        \textbf{BLEU} & 21.37 \\
        \textbf{ROUGE-L} & 41.86 \\
        \textbf{METEOR} & 39.06 \\
        \bottomrule
    \end{tabular}
    \caption{Example of feedback evaluation, for both ground-truth feedback and feedback generated by $\text{kNN}^{\text{one}}$ prompting. Generated predictions for both reference-free metrics are shown (with generated reasoning steps removed for brevity). The ground-truth feedback message is high quality and passes both reference-free metrics. However, the generated feedback message contains a misleading error, and fails the first reference-free metric while passing the second.}
    \label{table:feedback-eval-success-adj-fail}
\end{table*}

\begin{table*}[hb]
    \begin{tabular}{p{4.5cm}p{9cm}}
        \toprule
        \textbf{Question} & \[ 43.863 \] truncate this number to 1 decimal place \\
        \midrule
        \textbf{Correct Answer} & 43.8\\
        \midrule
        \textbf{Distractor} & 43.9\\
        \midrule
        \textbf{Ground-truth Feedback} & Not quite! Remember with truncating we just cut off the rest of the digits without any rounding.\\
        \textbf{Adjustment Prediction} & 43.8 \color{green}\ding{51}\\
        \textbf{Distractor Prediction} & 43.9 \color{green}\ding{51}\\
        \midrule
        \textbf{Generated Feedback ($\text{kNN}^{\text{one}}$)} & 43.863 truncated to 1 decimal place is 43.8.\\
        \textbf{Adjustment Prediction} & 43.8 \color{green}\ding{51}\\
        \textbf{Distractor Prediction} & 43.86 \color{red}\ding{55}\\
        \midrule
        \textbf{BLEU} & 0.0 \\
        \textbf{ROUGE-L} & 0.0 \\
        \textbf{METEOR} & 5.56 \\
        \bottomrule
    \end{tabular}
    \caption{Example of feedback evaluation, for both ground-truth feedback and feedback generated by $\text{kNN}^{\text{one}}$ prompting. Generated predictions for both reference-free metrics are shown (with generated reasoning steps removed for brevity). The ground-truth feedback message is high quality and passes both reference-free metrics. However, the generated feedback message simply gives away the answer without pointing out the error, and thus passes the first reference-free metric but fails the second.}
    \label{table:feedback-eval-success-dp-fail}
\end{table*}

\begin{table*}[hb]
    \begin{tabular}{p{4.5cm}p{9cm}}
        \toprule
        \textbf{Question} & \[ - (25)^{2} = \square \] \\
        \midrule
        \textbf{Correct Answer} & -625\\
        \midrule
        \textbf{Distractor} & 625\\
        \midrule
        \textbf{Ground-truth Feedback} & Not quite. As the negative is outside the bracket you need to work out 25 squared and then make the answer negative\\
        \textbf{Adjustment Prediction} & -625 \color{green}\ding{51}\\
        \textbf{Distractor Prediction} & 625 \color{green}\ding{51}\\
        \midrule
        \textbf{Generated Feedback (Zero-shot)} & Remember to apply the order of operations correctly. The negative sign should be applied first, then the exponent.\\
        \textbf{Adjustment Prediction} & -625 \color{green}\ding{51}\\
        \textbf{Distractor Prediction} & 625 \color{green}\ding{51}\\
        \midrule
        \textbf{BLEU} & 0.0 \\
        \textbf{ROUGE-L} & 20.0 \\
        \textbf{METEOR} & 15.35 \\
        \bottomrule
    \end{tabular}
    \caption{Example of feedback evaluation, for both ground-truth feedback and feedback generated by Zero-shot prompting. Generated predictions for both reference-free metrics are shown (with generated reasoning steps removed for brevity). The ground-truth feedback message is high quality and passes both reference-free metrics. However, the generated feedback states the incorrect order of operations, yet is able to elicit correct predictions for the reference-free metrics.}
    \label{table:feedback-eval-fail}
\end{table*}

\clearpage

\clearpage
\onecolumn

\subsection{Distractor Output Examples}

\begin{longtable}{p{4.5cm}p{9cm}}
    \toprule
    \textbf{Question} & Sheila is listing the multiples of 7 in ascending order. She starts with 7. What number should she write next? \\
    \midrule
    \textbf{Correct Answer} & 14\\
    \midrule
    \textbf{Authored Distractors} & $8$ $\quad$ $49$ $\quad$ $1$ $\quad$\\
    \midrule
    \textbf{Generated Distractors} & $7$ $\quad$ $21$ $\quad$ $28$ $\quad$\\
    \bottomrule
    \caption{Example of generated distractor quality in a ``failure'' case. No distractors overlap between human-authored and generated but all are reasonable distractor options. Distractors were generated by $\text{kNN}^{\text{all}}$ prompting strategy.}
    \label{table:distractor-example-mulitples} \\
\end{longtable}

\begin{longtable}{p{4.5cm}p{9cm}}
    \toprule
    \textbf{Question} & Convert $0.6$ to a fraction in its simplest form. \\
    \midrule
    \textbf{Correct Answer} & $\frac{3}{5}$\\
    \midrule
    \textbf{Authored Distractors} & $\frac{6}{10}$ $\quad$ $\frac{60}{100}$ $\quad$ $\frac{1}{6}$ $\quad$\\
    \midrule
    \textbf{Generated Distractors} & $\frac{6}{10}$ $\quad$ $\frac{5}{3}$ $\quad$ $\frac{10}{6}$ $\quad$\\
    \bottomrule
    \caption{Example of generated distractors where the key distractor is generated and the other distractors are unlikely to be selected. In this case the $\frac{6}{10}$ is selected by 28\% of students while the others are selected by less than 8\%. Distractors were generated by $\text{kNN}^{\text{all}}$ prompting strategy. }
    \label{table:distractor-example-fraction-simplify} \\
\end{longtable}

\clearpage
\clearpage

\subsection{Retrieved Examples for In-context Learning}
\label{sec:retrieved-examples}

\begin{minipage}{0.9\textwidth}
    \strut\newline
    \begin{tabular}{p{4cm}p{9cm}}
        \toprule
         \multirow{2}{*}{\textbf{Question}} & which multiplier can be used to find the value after an amount has decreased in value by 8\% for 4 years? \\
        & A) $\times 0.92^{4}$ $\quad$ B) $\times 0.08^{4}$ $\quad$ C)$\times 1.08^{4}$ $\quad$ D) $0.92 \times 4$ \\
        \midrule
         \multirow{2}{*}{\textbf{Example 1}} & which multiplier can be used to find the value after an amount has decreased in value by 5\% for 5 years? \\
        & A) $ \times 0.05^{5}$ $\quad$ B) $0.95 \times 5$ $\quad$ C) $\times 0.95^{5}$ $\quad$ D) $\times 1.05^{5}$ \vspace{10pt}\\        
         \multirow{2}{*}{\textbf{Example 2}} & the value of a laptop that initially cost \$1100, declines in value by 15\% a year. if you wanted to calculate the value of the tablet at the end of 6 years, what number would replace the square below? \[1100 \times \square^{6} \]\\
        & A) $3500 \times 0.85^{3}$ \ B) $3500 \times 1.5^{3}$ \ C) $3500 \times 0.85 \times 3$ \ D) $3500 \times 0.15^{3}$ \vspace{10pt}\\      
         \multirow{2}{*}{\textbf{Example 3}} & a car depreciates in value by 15\% each year. if a car was bought for \$3500, which of the following calculations would find the new value of the car after 3 years?\\
        & A) $5$ $\quad$ B) $4$ $\quad$ C) $6$ $\quad$ D) $8$ \\ 
        \bottomrule
    \end{tabular}
    \captionof{table}{Three examples retrieved by $\text{kNN}^{\text{all}}$.}
    \label{table:distractor-kNN-SEKFD}
\end{minipage}

\begin{table*}[hb]
    \begin{tabular}{p{4cm}p{9cm}}
        \toprule
         \multirow{2}{*}{\textbf{Question}} & which multiplier can be used to find the value after an amount has decreased in value by 8\% for 4 years? \\
        & A) $\times 0.92^{4}$ $\quad$ B) $\times 0.08^{4}$ $\quad$ C)$\times 1.08^{4}$ $\quad$ D) $0.92 \times 4$ \\
        \midrule
         \multirow{2}{*}{\textbf{Example 1}} & $(-\frac{1}{4})^{-2} \equiv \square$\\
        & A) $-\frac{1}{16}$ $\quad$ B) $16$ $\quad$ C)$\frac{1}{16}$ $\quad$ D) $-16$ \vspace{10pt}\\
         \multirow{2}{*}{\textbf{Example 2}} & what is the correct multiplier for decreasing by 30\%?\\
        & A) $\times 1.3$ $\quad$ B) $\times 0.7$ $\quad$ C) $\times 30$ $\quad$ D) $\times 0.3$\vspace{10pt} \\        
         \multirow{2}{*}{\textbf{Example 3}} & dividing by 5 and then by 4 is the same as dividing by ...\\
        & A) $1.25$ $\quad$ B) $20$ $\quad$ C) $1$ $\quad$ D) $0.8$ \\          
        \bottomrule
    \end{tabular}
    \caption{Three examples retrieved by Random.}
    \label{table:distractor-random}
\end{table*}

\begin{table*}[hb]
    \begin{tabular}{p{4cm}p{9cm}}
        \toprule
         \multirow{2}{*}{\textbf{Question}} & which multiplier can be used to find the value after an amount has decreased in value by 8\% for 4 years? \\
        & A) $\times 0.92^{4}$ $\quad$ B) $\times 0.08^{4}$ $\quad$ C)$\times 1.08^{4}$ $\quad$ D) $0.92 \times 4$ \\
        \midrule
         \multirow{2}{*}{\textbf{Example 1}} & which is the correct multiplier for decreasing by 40\%?\\
        & A) $\times 1.4$ $\quad$ B) $\times 40$ $\quad$ C) $\times 0.6$ $\quad$ D) $\times 0.4$ \vspace{10pt}\\          
         \multirow{2}{*}{\textbf{Example 2}} & a company bought a van for \$12000. the van lost $\frac{1}{10}$ of its value each year. after how many years was the value of the van worth \$6377.29?\\
        & A) $5$ $\quad$ B) $7$ $\quad$ C) $6$ $\quad$ D) $8$ \vspace{10pt}\\     
         \multirow{2}{*}{\textbf{Example 3}} & a company bought a van for \$15000. the van lost $\frac{1}{10}$ of its value each year. After how many years was the value of the van worth \$8857.29?\\
        & A) $5$ $\quad$ B) $4$ $\quad$ C) $6$ $\quad$ D) $8$ \\   
        \bottomrule
    \end{tabular}
    \caption{Three examples retrieved by $\text{kNN}^{\text{all}}_{\neg T}$.}
    \label{table:distractor-kNN-SEKFD*}
\end{table*}

\end{document}